\documentclass{article}

\usepackage[preprint,nonatbib]{neurips_2024}

\usepackage{enumitem}
\usepackage[utf8]{inputenc} % allow utf-8 input
\usepackage[T1]{fontenc}    % use 8-bit T1 fonts
\usepackage{hyperref}       % hyperlinks
\usepackage{url}            % simple URL typesetting
\usepackage{booktabs}       % professional-quality tables
\usepackage{amsfonts}       % blackboard math symbols
\usepackage{nicefrac}       % compact symbols for 1/2, etc.
\usepackage{microtype}      % microtypography
\usepackage{xcolor}         % colors
\usepackage{graphicx}
\usepackage{subfig}

\usepackage[square, numbers]{natbib}
\setcitestyle{numbers} % 设置 author-year 引用样式

\definecolor{mypurple}{HTML}{806491}
\definecolor{myorange}{HTML}{F5704A}
\definecolor{my_awai_aoi}{HTML}{2F70AF}
\definecolor{my_awai_gray}{HTML}{C0C0C0}
\definecolor{mistral}{HTML}{ff7000}

\usepackage{pgf}
\usepackage{multirow}
\usepackage{tikz}
\usepackage{pgfplots}
\pgfplotsset{compat=1.16}
\usetikzlibrary{patterns}
\usepackage{colortbl}
\usepackage{scalefnt} %用于设置字体
\usepackage{bm}
\usepackage{tcolorbox}
\tcbuselibrary{most}
\tcbuselibrary{xparse}
\usepackage{epstopdf}
\usepackage{caption}
\usepackage{subcaption}
\usepackage{tkz-kiviat}
\usetikzlibrary{arrows}
\usepackage{wrapfig}
\usepackage{algorithm2e}
\definecolor{color1}{RGB}{145,30,180}
\definecolor{color2}{RGB}{245,130,48}
\definecolor{color3}{RGB}{230,25,75}
\definecolor{my_dark_midori}{HTML}{455442}

\usepackage{collcell}
\usepackage{colortbl,dcolumn}
\definecolor{my_awai_red}{HTML}{B9848C}
\newcommand{\ColorGradient}[5]{ % #1=value, #2=Max, #3=Min, #4=color #5=text
\pgfmathparse{max(min(100.0*(#1 - #3)/(#2-#3),100.0),0.00)}\xdef\PercentColor{\pgfmathresult}
\cellcolor{#4!\PercentColor!white}#5
}

\newcommand*{\MinNumberOne}{45.3}%
\newcommand*{\MaxNumberOne}{56.7}%
\newcommand{\ApplyGradientRedOne}[1]{%
        \ifnumber #1
        \pgfmathsetmacro{\PercentColor}{max(min(100.0*(#1 - \MinNumberOne)/(\MaxNumberOne-\MinNumberOne),100.0),0.00)} %
        \hspace{-0.33em}\colorbox{my_awai_red!\PercentColor!white}{#1}
        \fi
}
\newcolumntype{R}{>{\collectcell\ApplyGradientRedOne}c<{\endcollectcell}}

\newcommand*{\MinNumberTwo}{29.0}%
\newcommand*{\MaxNumberTwo}{63.6}%
\newcommand{\ApplyGradientAoiTwo}[1]{%
        \ifnumber #1
        \pgfmathsetmacro{\PercentColor}{max(min(100.0*(#1 - \MinNumberTwo)/(\MaxNumberTwo-\MinNumberTwo),100.0),0.00)} %
        \hspace{-0.33em}\colorbox{my_awai_aoi!\PercentColor!white}{#1}
        \fi
}
\newcolumntype{A}{>{\collectcell\ApplyGradientAoiTwo}c<{\endcollectcell}}

\newcommand*{\MinNumberThree}{66.4}%
\newcommand*{\MaxNumberThree}{85.8}%
\newcommand{\ApplyGradientRedThree}[1]{%
        \ifnumber #1
        \pgfmathsetmacro{\PercentColor}{max(min(100.0*(#1 - \MinNumberThree)/(\MaxNumberThree-\MinNumberThree),100.0),0.00)} %
        \hspace{-0.33em}\colorbox{my_awai_red!\PercentColor!white}{#1}
        \fi
}
\newcolumntype{S}{>{\collectcell\ApplyGradientRedThree}c<{\endcollectcell}}

\newcommand*{\MinNumberFour}{50.8}%
\newcommand*{\MaxNumberFour}{80}%
\newcommand{\ApplyGradientAoiFour}[1]{%
        \ifnumber #1
        \pgfmathsetmacro{\PercentColor}{max(min(100.0*(#1 - \MinNumberFour)/(\MaxNumberFour-\MinNumberFour),100.0),0.00)} %
        \hspace{-0.33em}\colorbox{my_awai_aoi!\PercentColor!white}{#1}
        \fi
}
\newcolumntype{B}{>{\collectcell\ApplyGradientAoiFour}c<{\endcollectcell}}

\newcommand*{\MinNumberMM}{67.8}%
\newcommand*{\MaxNumberMM}{71}%
\newcommand{\ApplyGradientRedMM}[1]{%
        \pgfmathsetmacro{\PercentColor}{max(min(100.0*(#1 - \MinNumberMM)/(\MaxNumberMM-\MinNumberMM),100.0),0.00)} %
        \hspace{-0.33em}\colorbox{my_awai_red!\PercentColor!white}{#1}
}
\newcolumntype{M}{>{\collectcell\ApplyGradientRedMM}c<{\endcollectcell}}

\newcommand*{\MinNumberGG}{78.8}%
\newcommand*{\MaxNumberGG}{85}%
\newcommand{\ApplyGradientAoiGG}[1]{%
        \pgfmathsetmacro{\PercentColor}{max(min(100.0*(#1 - \MinNumberGG)/(\MaxNumberGG-\MinNumberGG),100.0),0.00)} %
        \hspace{-0.33em}\colorbox{my_awai_aoi!\PercentColor!white}{#1}
}
\newcolumntype{N}{>{\collectcell\ApplyGradientAoiGG}c<{\endcollectcell}}

\newcommand*{\MinNumberAA}{84.6}%
\newcommand*{\MaxNumberAA}{89}%
\newcommand{\ApplyGradientRedAA}[1]{%
        \pgfmathsetmacro{\PercentColor}{max(min(100.0*(#1 - \MinNumberAA)/(\MaxNumberAA-\MinNumberAA),100.0),0.00)} %
        \hspace{-0.33em}\colorbox{my_awai_red!\PercentColor!white}{#1}
}
\newcolumntype{O}{>{\collectcell\ApplyGradientRedAA}c<{\endcollectcell}}

\title{Brainstorming Brings Power to Large Language Models of Knowledge Reasoning}

\author{%
  Zining Qin, Chenhao Wang, Huiling Qin*, Weijia Jia \\
  Beijing Nomal University\\
 \texttt{orekinana@gmail.com} \\
  \AND
\And
  \And
}

\begin{document}

\maketitle

\begin{abstract}
Large Language Models (LLMs) have demonstrated amazing capabilities in language generation, text comprehension, and knowledge reasoning. While a single powerful model can already handle multiple tasks, relying on a single perspective can lead to biased and unstable results. Recent studies have further improved the model's reasoning ability on a wide range of tasks by introducing multi-model collaboration. However, models with different capabilities may produce conflicting answers on the same problem, and how to reasonably obtain the correct answer from multiple candidate models has become a challenging problem. In this paper, we propose the multi-model brainstorming based on prompt. It incorporates different models into a group for brainstorming, and after multiple rounds of reasoning elaboration and re-inference, a consensus answer is reached within the group. We conducted experiments on three different types of datasets, and demonstrate that the brainstorming can significantly improve the effectiveness in logical reasoning and fact extraction. Furthermore, we find that two small-parameter models can achieve accuracy approximating that of larger-parameter models through brainstorming, which provides a new solution for distributed deployment of LLMs.
\end{abstract}

\section{Introduction}

With the accumulation of data and the significant increase in computational power, large language models (LLMs) rely on their vast training data and scale to show excellent capabilities in language understanding, text generation, and knowledge reasoning\cite{brown2020language,bubeck2023sparks,ouyang2022training}. Although existing LLMs can solve many different types of tasks, the reasoning bias caused by the limitations and instability of a single model's perspective still cannot be ignored. Recent research indicates that fusing multiple models to work collaboratively can further enhance reasoning capabilities on a wide range of tasks \cite{cai2023self,jiang2023llm,li2024camel,liang2023encouraging,thoppilan2022lamda,wang2023fusing,xiong2023examining}. For example, one promising approach is to incorporate multiple agents generated by an LLM into a group  after substituting different roles and designing a special interaction mechanism \cite{du2023improving}, and it can improve the diversity of reasoning while allowing agents to provide more stable and reliable results through debates in the interactions \cite{chan2023chateval}.

\begin{figure}[htbp]
    \centering
    \subfloat[Vote]{\includegraphics[width=0.32\textwidth]{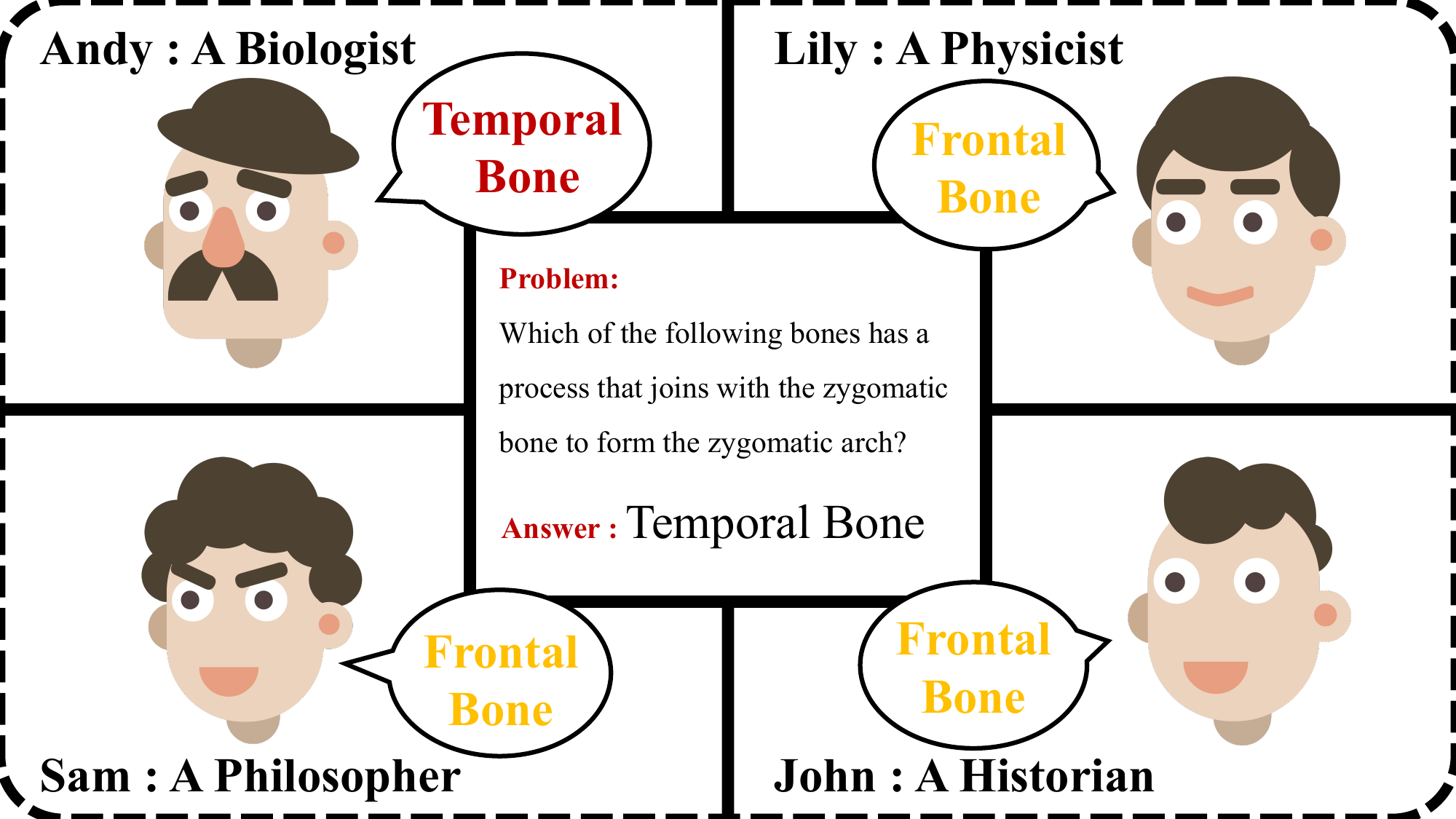}}
    \hfill % adds some space between the subfigures
    \subfloat[Review]{\includegraphics[width=0.32\textwidth]{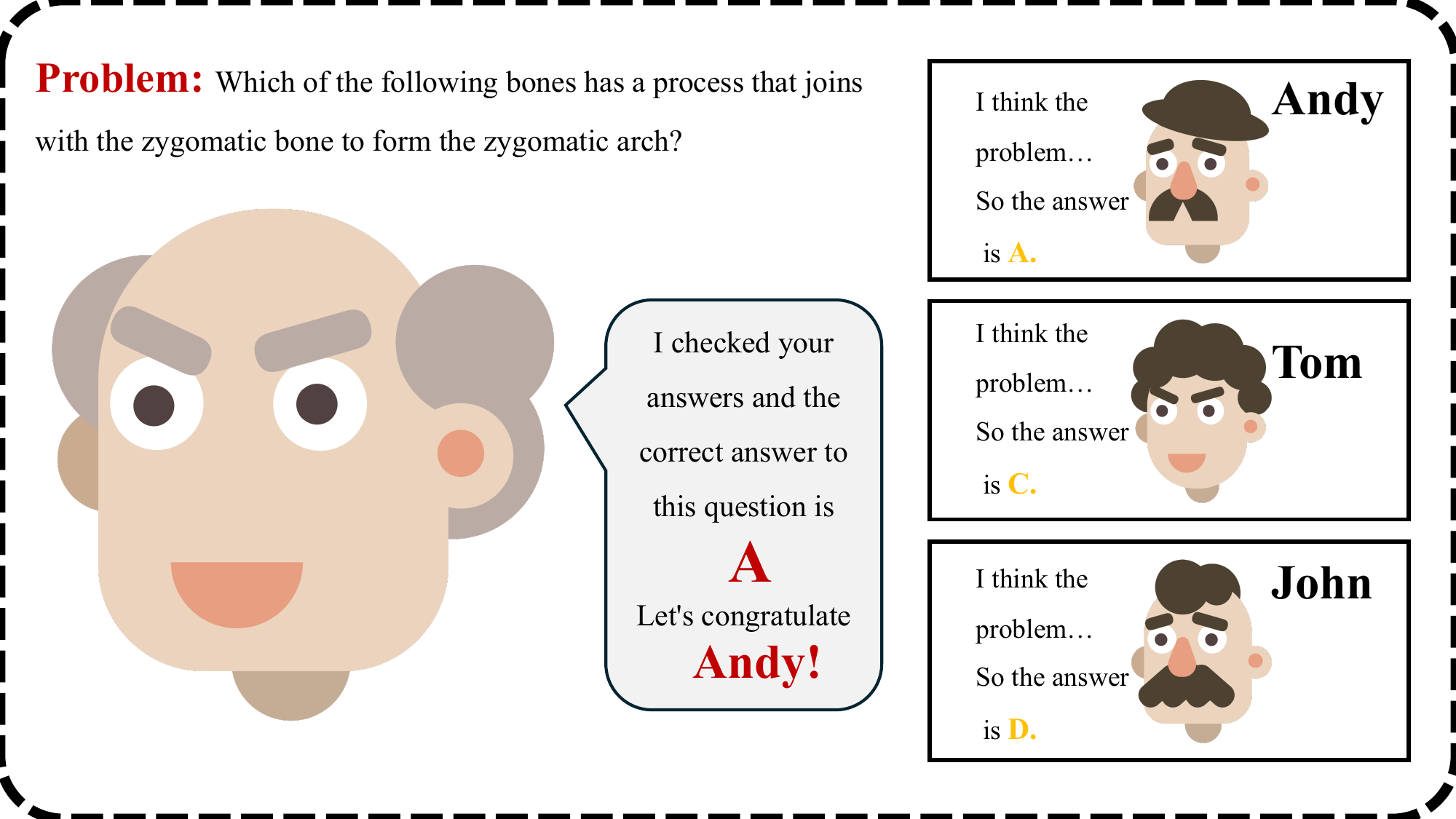}}
    \hfill % adds more space between the subfigures
    \subfloat[Brainstorm]{\includegraphics[width=0.32\textwidth]{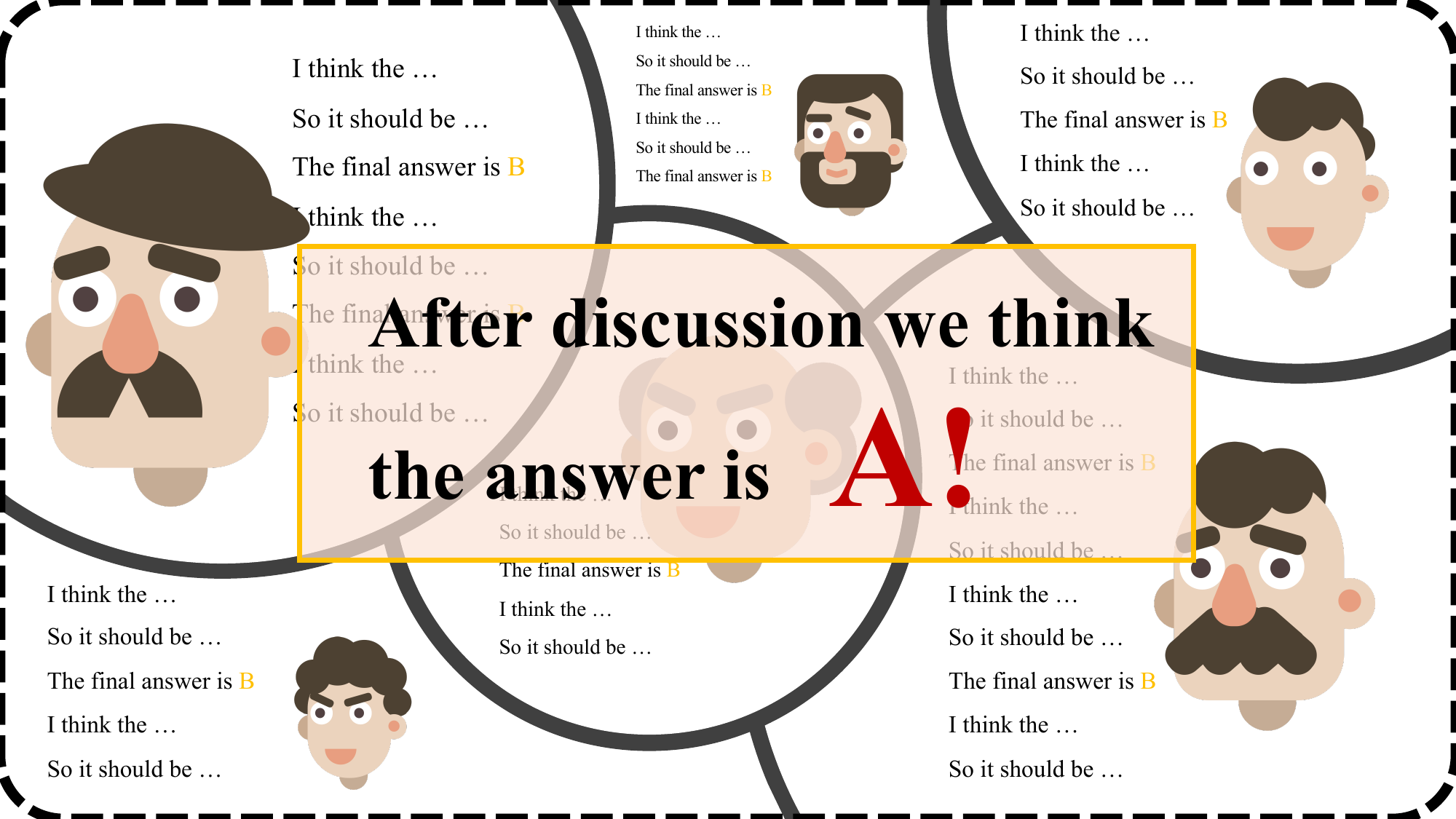}}
    \caption{Differences in getting answers through voting, strong model judgments (review), and brainstorming of many different types of collaboration on large language models. }
    \label{fig:intro}
  \end{figure}

However, in the knowledge reasoning scenario that emphasizes the availability of definitive answers, the roles played by agents for the same type of scientific questions tend to be uniform. In such a case, the reasoning perspective of agents based on the single %same
model becomes narrow, necessitating the involvement of models with different capabilities to improve the overall reasoning performance. %However, 
When multiple models reasoning, due to the differences in the training data and training process, models with different capabilities may provide different answers to the same question. Therefore, in the face of conflicting results of reasoning, it becomes a critical and challenging problem to formulate a rule to ensure that the group provides a uniform answer from multiple candidates. Figure \ref{fig:intro} shows three natural interaction methods to obtain the final answer.

The most straightforward approach is voting, as shown in Figure \ref{fig:intro} (a). Multiple models provide what they consider to be the correct answer, and the final answer is then determined according to the majority rule after counting the number of votes for each candidate answer \cite{li2024more,yang2024llm}. It is noticed that voting-based approaches tend to lack interpretability, and if the ability of a few models is significantly stronger than that of the majority, then the result obtained by voting may miss the correct answer. A further step can be taken by introducing an additional discriminative model as a reviewer \cite{shinn2024reflexion} that evaluates multiple answers and then gives the final answer, shown as Figure \ref{fig:intro} (b). However, the reviewer requires a higher level of knowledge and judgment than models in the group, and it is very difficult to construct such a discriminative model.

Drawing inspiration from the collective thinking and reasoning process of human beings \cite{street2024llm}, we propose a prompt-based multi-model brainstorming approach, %namely \emph{LLM Brainstorming}, 
as shown in Figure \ref{fig:intro} (c). It allows each model to substitute a similar expert role and iteratively incorporate the reasoning process of other models into their thinking scope to update their answers, until a consensus is reached.
Furthermore, in order to enhance the diversity of group thinking, we select the model with some differences in performance on different tasks for brainstorming. This ensures that each model is capable of bringing novel ways of thinking to knowledge reasoning. In this way, models enable thinking from different perspectives and capture issues that are difficult to identify from a single viewpoint.

The main contributions of our work are stated as follows:
\begin{itemize}[leftmargin=*]
  \item We propose a prompt-based multi-model brainstorming that has a more nuanced mindset than single model and the extended multi-agent debate approaches, which enhances the reasoning accuracy in both logical and factual terms.
  \item We utilize multi-model brainstorming instead of manual prompts in Chain of Thought (CoT)\cite{kojima2022large,wei2022chain} to reduce the cost of manual labeling. We use two small models with brainstorming to replace one large model, and exhibit approximate, even higher accuracy in knowledge reasoning in diversified datasets, providing a new solution for distributed deployment of large models. 
   % We use two small models with brainstorming to replace one large model with guaranteed inference accuracy, providing a new solution for distributed deployment of large models. \bluecomment{Precisely, over the diversified datasets examined by our experiments, the brainstorm of  two small models exhibit considerably higher accuracy in knowledge reasoning than one large model with double parameter size. }
  \item We find that the LLM is able to integrate different viewpoints in the brainstorm. The redundant historical dialogues or an additional summary are unnecessary to the reasoning, which avoids the problem of reduced reasoning ability due to excessive space occupation in multiple generation.
\end{itemize}

\section{Related Work}
% Related works can be organized as following parts: 1) LLM self-ensemble (Wang et al., 2023b), which attempts to harness multiple outputs from homogeneous LLMs to assemble the final answer; 

\textbf{Prompts Enhancement in LLM Reasoning.} Due to the versatile capabilities that LLMs demonstrate, researchers have leveraged LLMs as the foundation to build AI agents that can communicate with human and other LLMs and have achieved significant progress \cite{xi2023rise}, including improvements via various prompts\cite{li2022advance,jung2022maieutic,wang2022iteratively,xiong2023examining,zhang2022automatic}.
% CoT, zs-cot, stepback, Solo Performance Prompting (SPP)
Few shot and zero shot Chain-of-thought (CoT) \cite{kojima2022large,wei2022chain} encourage LLMs to analyze the input question step by step following the procedure given by human in reasoning and generation, making the responses readable and interpretable to human and other LLM agents. 
Wang et al. \cite{wang2023unleashing} propose self-collaboration method to enable the communication of agents by assigning multi-persona to a single LLM via prompts.
Zheng et al. \cite{zheng2023take} enable LLMs to ``recall'' and derive high-level concepts and first principles from question instances by applying Step-back prompting. With additional hints respect to the question generated by LLM agent itself, the agent gain more chances of utilizing such principles properly and answering correctly. Our work also adopts some prompting techniques to enhance the quality of LLMs' responses.
% Liet al. (2023b); Zhang et al. (2023) also propose a similar approach. However, they probe different dimensions of improving LLM-based evaluators and do not explore the effectiveness of natural language interaction.

\textbf{Interaction Framework for Multiple LLMs.}
Compared to models that only take vectors specifically designed for them as input, LLMs bridge the gap between models and human using natural language via their tokenizers. Thanks to this property researchers can try various frameworks and strategies to combine multiple LLMs for downstream tasks.
Mixture of Experts \cite{jiang2024mixtral} trains a router network to select a subset composed of parts of LLMs (experts) to process the inner states and combine their outputs.
% more agent voting, debate, chateval, teacherstudent
Li et al. \cite{li2024more} conduct a series of experiments on majority voting among large numbers of LLMs and the performance of voting relies considerably on the great number of agents.
Li et al. \cite{li2024camel} and Chen et al. \cite{chen2023agentverse} propose a cooperative framework facilitating LLMs' autonomous cooperation to solve complex tasks.
% Park et al. (2023) create a sandbox environment consisting of 25 individual virtual entities endowed with a character description and memory system. Every intelligent agent is capable of autonomously interacting with other agents and the environment simulating reliable human behavior. 
% Dong et al. (2023); Qian et al. (2023) incorporate a waterfall model to manage a multi-agent virtual team towards software development. 
% Liu et al. (2023a) utilize a sandbox environment to curate reliable datasets in better alignment with human preference and train a socially-aligned LLM. 
% Mandi et al. (2023) propose a novel framework designed for the collaboration of multiple robots, utilizing multiple LLMs to enhance coordination and strategic planning among the robots. 
Chan et al.\cite{chan2023chateval} build a multi-agent referee team called ChatEval to autonomously discuss and evaluate the quality of different texts, inspired by best practices of human evaluation processes.
Concurrent with our work, Liang et al. \cite{liang2023encouraging} and Du et al. \cite{du2023improving} also propose a similar method, making use of the multi-agent debate framework in translation, arithmetic and some optimization problems. % and resulting in better performance. 
However, Liang et al. \cite{liang2023encouraging} focus on the diversity that emerges from debate rather than the reasoning accuracy, while Du et al. \cite{du2023improving} insufficiently explore the effectiveness and efficiency of the natural language interaction among LLMs.

\begin{figure*}[htbp]
    \centering
    \includegraphics[width=\textwidth]{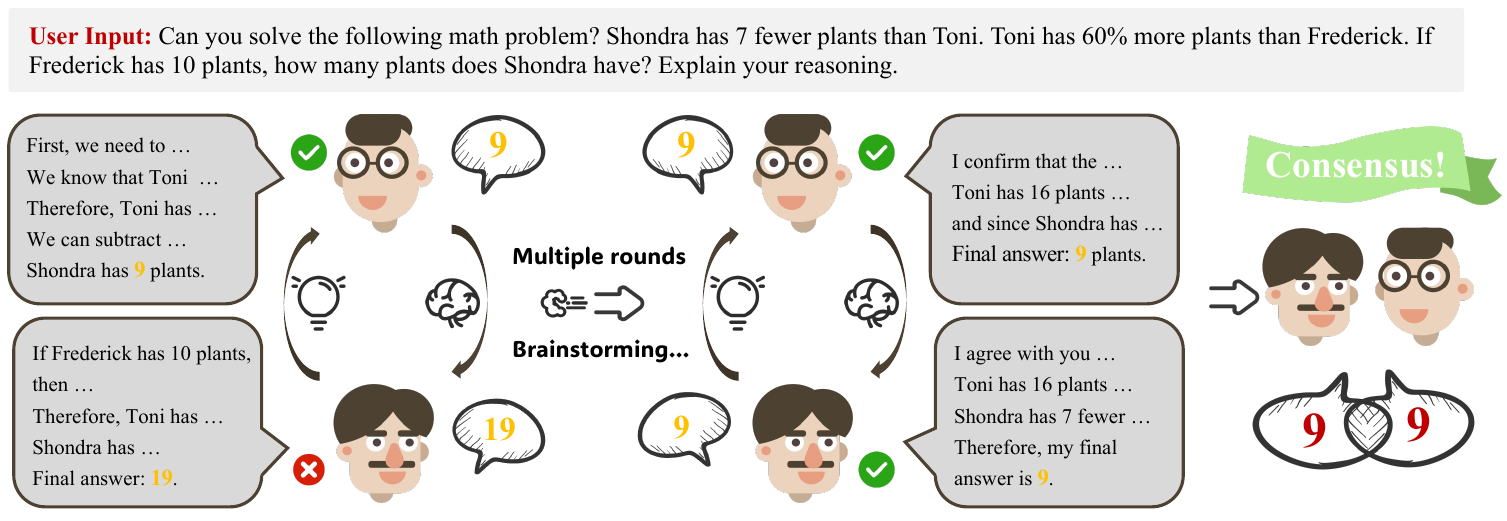}
    \caption{An illustration of LLM Brainstorming. When multiple models are involved in the inference process, they can discuss with each other and eventually reach a consensus judgment.}
    \label{fig:framework}
\end{figure*}
\section{Methedology}
In this section, we elaborate on the principal components in LLM Brainstorming including Multi-model brainstorming, consensus in brainstorming, dialog truncating strategy, and provide a detailed overview of each component’s role and functionality.

\textbf{Multi-model Brainstorming.} Using multiple heterogeneous LLMs for brainstorming is the key to our method. The reasoning performance of LLMs in various domains is closely related to the corpora used during training. Different organizations are likely to utilize closed-source datasets built by themselves in addition to open-source corpora for training LLMs, resulting in models' performance varying across some domains. Therefore, utilizing different models can provide more wealth of information during brainstorming. In the first round of the brainstorming, we give the question to models without additional contents in prompt. If models response with different answers, we incorporate reasoning processes as thought samples from other models into the prompt for the next round. That is, a model receives responses from the others and, in turn, sends its own responses to them in each round of the brainstorming. The brainstorming process is shown as Figure \ref{fig:framework}. With other models' reasoning output embedded, responses of models to questions can be regarded as a specific CoT (Chain-of-Thought) with more information related to the question, which can lead models to the correct answer.  Note that constructing prompts for each model in brainstorming on any question does not require human intervention. The complete prompt templates are in Appendix \ref{sec:prompt}.
% brainstorm 过程的描述有点模糊，很多句子的主语不知道是在说哪个模型，梳理一下语言逻辑再修改一下
%LLMs are trained with multi-source corpora and obtain various knowledge; LoRa enhances LLM in some domain. Although the domains that LLMs perform well are different, logical ability are general for many questions. Besides, for memory questions...
% different responses to one question can be regarded as a specific CoT, which may assist the LLM reasoning further based on current analysis and lead LLMs to the true answer...

\textbf{Consensus in Brainstorming.} The brainstorming process reveals multiple reasoning pathways and answers from LLMs' responses. How can we get the final  answer with or without consensus? \emph{Consensus} or \emph{convergence} in LLM brainstorming occurs when all LLMs  provide the same answer in the same round of brainstorming, and the final answer is this same answer. Since there is no guarantee of consensus under all conditions, we adopt a parameter named \verb|max_brainstorming_round| to control the brainstorming. If not all models in brainstorming provide the same answer (such as choices and numerical values), they will start another round of brainstorming until consensus is achieved or the number of brainstorming rounds reaches the \verb|max_brainstorming_round|. As there is no confidence estimation for brainstorming, we treat all the answers generated by all the LLMs throughout the brainstorming process equally without introducing extra bias.  When the brainstorming ends without consensus, we record all the answers generated by all the LLMs throughout the brainstorming process and returns the most frequent answer as the final result. This approach ensures that the final answer reflects the collective reasoning and most frequently agreed-upon responses of all the models over multiple rounds. The dialogs of LLM brainstorming are readable to human and we can easily catch up with the reasoning of brainstorming to the final answers.
% Aggregation是什么操作？？the most common answer是什么，怎么产生的，为什么没收敛要这么取最终结果，converge 取一致答案是很自然的，但是把历史回答拿来 vote 取最终答案不是很直接，补充解释一下为什么要这么做
% Control of max Brainstorming rounds....(Brainstorming is more readable than ensemble vote)

\textbf{Dialog Truncation Strategy.} Since our method involves multiple rounds of conversation, the context will contain considerable redundancy. To ensure the efficiency and accuracy of the reasoning, we adopt dialog truncation strategies. Besides the first round of brainstorming, which includes the user's question and the reasoning and answers generated in the first round, we only keep the last $n$ rounds of brainstorming dialogs as input for the next round. Different parameter values of $n$ represent different dialog truncation strategies. We note that although LLMs can capture the entire brainstorming without truncation, the linear growth of the length of input tokens and the time consumption of reasoning may lead to unacceptable costs. In the experiments, we set $n=1$ because duplicate reasoning segments are common as the number of rounds increases, which can confuse LLMs and undermine the potential for correct reasoning. In addition, LLMs are good at summarizing previous answers, even when there is no summary order in the prompts. Truncating middle parts of dialogs can reduce input tokens and speed up brainstorming without losing the stem of previous reasoning and the question. 
% 可以直接说我们在这里 n 取 1，因为对于同一问题多轮对话有较大冗余，考虑到在 prompt 中让大模型仔细思考+大模型有一定的总结能力，所以我们直接取最新一轮来概括历史对话的推理和结论
% Providing the whole brainstorming dialog may help LLMs have command of all the details, while the cost of reasoning or calculation growing multiplicatively. As we find that the analysis in brainstorming will goes similar and cannot provide enough useful information, we try several dialog truncating strategies. ....

\section{Experiments}
We evaluate the LLM %multi-model 
brainstorming on three datasets from different domains, which include the logical and factual reasoning question answer. For the LLMs in brainstorming, we use ChatGPT \cite{achiam2023gpt} and Gemini Pro \cite{reid2024gemini,team2023gemini} considering their strong capability shown in the literature and practice, and we also use some smaller open-sourced models such as Qianwen1.5-7B \cite{bai2023qwen}, Baichuan2-7B \cite{yang2023baichuan} and Mistral-7B \cite{jiang2024mixtral}(Qwen, Baichuan, Mistral in short, respectively). The experimental evaluation is conducted in terms of accuracy, efficiency, and characteristic of the brainstorming.

\subsection{Datasets}
In order to evaluate the effectiveness of the brainstorming by extensive tasks, we test the ability of LLMs with respect to different factual knowledge questions. We utilize the existing MMLU dataset \cite{hendrycks2020measuring} to benchmark the accuracy of responses, which covers questions from subdomains Math, Social, Business, Humanities, Medical, etc. To further evaluate the reasoning ability in logic tasks, we consider harder mathematical reasoning tasks. Using GSM dataset \cite{cobbe2021training}, the models need to correctly solve grade school mathematical reasoning tasks. To evaluate the improvement of brainstorming in factual reasoning, we consider two tasks with different levels of difficulty in factual knowledge. We use the ARC-easy and ARC-challenge datasets \cite{clark2018think}, which contain only natural, grade-school science questions, to test the model's memory and ability to capture key knowledge.

\subsection{Baselines}
We evaluate the multi-model brainstorming by the following methods. {By default, the multi-model brainstorming involves a pair of LLMs, and at most 8 rounds of interactions are implemented no matter whether a consensus is reached or not.} As the effectiveness  comparison, we {compare the multi-model brainstorming with}  the \textbf{\textit{single-model-based}} method, {the \textbf{\emph{ensemble vote}}}, and  the \textbf{\textit{multi-agent-based}} method in the evaluation experiment. %Single-model means that
{In the single-model-based method,} we directly query an LLM to generate the response towards the evaluation. %Multi-agent means that 
{In the ensemble voter, the answer is determined by the majority rule of multiple models. In the multi-agent-based method,} we employ multiple agents from the same LLM, but not multiple different LLMs, in the interaction process to obtain the answer. 
%By default, we configure the communication strategy to one-by-one, model numbers to 2, and brainstorming rounds upper to 8. 
{For all methods, the models are three open-source LLMs with diverse performance in different domains, Qwen-7B \cite{bai2023qwen}, Baichuan-7B \cite{yang2023baichuan}, Mistral-7B \cite{jiang2024mixtral}, and two larger models, GPT-3.5 \cite{achiam2023gpt} and Gemini Pro \cite{reid2024gemini,team2023gemini}. }
%The participant models are selected from three open-source LLM with diverse performance in different domains, as well as two larger models, including Qianwen-7B \cite{bai2023qwen}, Baichuan-7B \cite{yang2023baichuan}, Mistral-7B \cite{jiang2024mixtral}, and GPT-3.5 \cite{achiam2023gpt}, Gemini Pro \cite{reid2024gemini,team2023gemini}. 

For the evaluation of efficiency on the question answer task, we compare the brainstorming with \textbf{\textit{models using CoT prompt}} and \textbf{\textit{models with double-parameter}}. The CoT based models we utilize are Qianwen-7B \cite{bai2023qwen}, Baichuan-7B \cite{yang2023baichuan}, and Mistral-7B \cite{jiang2024mixtral} with 5-shot for MMLU, 8-shot for GSM, and zero-shot for ARC. In order to adapt to the 7B model size, we select Qianwen-14B  \cite{bai2023qwen} and Baichuan-13B \cite{yang2023baichuan} as the models with double parameter size. Since the Mixtral model does not have a version of size 14B, we use the MoE version (Mixtal 7$\times$8B, with 12.9B active parameters and 46.7B total parameters) \cite{jiang2024mixtral} instead.

\subsection{Results}\label{sec:result}

\paragraph{Accuracy Improvement for Reasoning. }

% Tablet 1: vsSelf
\begin{table}[htbp]
        \begin{center}
          \caption{The accuracy of open-source models (Baichuan, Qwen, 
          % \bluecomment{(the shorthand for Qianwen)}
          and Mistral) using brainstorming, single model, ensemble and multi-agent reasoning strategies on different datasets. 
          The scale of all the models is 7B. $\overline\Delta$ represents the average improvement compared to the related single-models.%%%%%%%%%%%%%%%%
          }
          % \textbf{The Accuracy of LLM Brainstorming and Baseline Methods over Multiple Tasks.} Our method outperforms Ensemble Vote and Multi-Agent across the four tasks. The deeper color in the column "Acc." means the higher accuracy on the relative task.  The bolded instances indicate the highest improvement on the relative task.
          \label{tab:acc_open_source}
          % \begin{tabular}{cr|cccc|}
          %     & & \multicolumn{4}{c|}{Qwen-7B $\times$ Baichuan2-7B}\\
          %   & & MMLU & GSM & ARC-e & ARC-c \\
          %   \hline
            
          %   \multicolumn{2}{c|}{Game} & 53.2 & 54.1 & 81.4 & 64.0 \\ 
          %   \arrayrulecolor{my_awai_gray}\hline
          %   & Qwen-7B & 54.8 & 58.6 & 80.1 & 66.4 \\
          %   Self (2$\times$) & Baichuan2-7B & 45.9 & 32.4 & 72.8 & 53.3 \\
          %   & Mistral-7B &  &  &  &  \\
          %   % 3 & 23.113231 & c\\
          %   % 4 & 25.113231 & d\\
          % \end{tabular}
      
          % \begin{tabular}{c|RASB|}
          \resizebox{\textwidth}{!}{
          \begin{tabular}{c|cccccccc}
              % & & \multicolumn{4}{c|}{Qwen-7B $\times$ Baichuan2-7B}\\
              % \makebox[0.2\textwidth][c]{Method} &
              \multirow{2}{*}{Method} &
              \multicolumn{2}{c}{MMLU} &
              % \makebox[0.1\textwidth][c]{MMLU(\%)} &
              \multicolumn{2}{c}{GSM} &
              % \makebox[0.1\textwidth][c]{GSM(\%)} &
              \multicolumn{2}{c}{ARC-e} &
              % \makebox[0.1\textwidth][c]{ARC-e(\%)} &
              \multicolumn{2}{c}{ARC-c} \\
              % \makebox[0.1\textwidth][c]{ARC-c(\%)} \\
            % Method & MMLU(\%) & GSM(\%) & ARC-e(\%) & ARC-c(\%) \\
           & \makebox[0.7cm][c]{Acc.}& \makebox[0.3cm][c]{$\overline\Delta$}&
             \makebox[0.7cm][c]{Acc.}& \makebox[0.3cm][c]{$\overline\Delta$}& 
             \makebox[0.7cm][c]{Acc.}& \makebox[0.3cm][c]{$\overline\Delta$}& 
             \makebox[0.7cm][c]{Acc.}& \makebox[0.3cm][c]{$\overline\Delta$}\\
            % \midrule
            \hline
            \textbf{Single-Model} \\
            Baichuan & 
            \ColorGradient{45.3}{56.7}{45.3}{my_awai_red}{45.3\%} & - &
            \ColorGradient{29.0}{63.6}{29.0}{my_awai_aoi}{29.0\%} & - &
            \ColorGradient{70.4}{85.8}{66.4}{my_awai_red}{70.4\%} & - &
            \ColorGradient{50.8}{80.0}{50.8}{my_awai_aoi}{50.8\%} & - \\
            Mistral & 
            \ColorGradient{46.2}{56.7}{45.3}{my_awai_red}{46.2\%} & - & 
            \ColorGradient{45.9}{63.6}{29.0}{my_awai_aoi}{45.9\%} & - & 
            \ColorGradient{70.6}{85.8}{66.4}{my_awai_red}{70.6\%} & - & 
            \ColorGradient{60.4}{80.0}{50.8}{my_awai_aoi}{60.4\%} & - \\
            Qwen & 
            \ColorGradient{49.3}{56.7}{45.3}{my_awai_red}{49.3\%} & - & 
            \ColorGradient{51.1}{63.6}{29.0}{my_awai_aoi}{51.1\%} & - & 
            \ColorGradient{66.4}{85.8}{66.4}{my_awai_red}{66.4\%} & - & 
            \ColorGradient{55.3}{80.0}{50.8}{my_awai_aoi}{55.3\%} & - \\
            % \arrayrulecolor{my_awai_gray}\bottomrule
            % \arrayrulecolor{my_awai_gray}
            \hline
            \textbf{Ensemble Vote} \\
            Qwen $\times$ Baichuan $\times$ Mistral & 
            \ColorGradient{54.6}{56.7}{45.3}{my_awai_red}{54.6\%} & +7.7\% & 
            \ColorGradient{54.9}{63.6}{29.0}{my_awai_aoi}{54.9\%} & +12.9\% & 
            \ColorGradient{79.2}{85.8}{66.4}{my_awai_red}{79.2\%} & +10.1\% & 
            \ColorGradient{64.9}{80.0}{50.8}{my_awai_aoi}{64.9\%} & +9.4\% \\
            \hline
            \textbf{Multi-Agent} \\
            % Self: 
            Baichuan$\times$2 & 
            \ColorGradient{45.9}{56.7}{45.3}{my_awai_red}{45.9\%} & +0.6\% & 
            \ColorGradient{32.4}{63.6}{29.0}{my_awai_aoi}{32.4\%} & +3.4\%& 
            \ColorGradient{72.8}{85.8}{66.4}{my_awai_red}{72.8\%} & +2.4\%& 
            \ColorGradient{53.3}{80.0}{50.8}{my_awai_aoi}{53.3\%} & +2.5\% \\
            % Self: 
            Mistral$\times$2 & 
            \ColorGradient{51.4}{56.7}{45.3}{my_awai_red}{51.4\%} & +5.2\%& 
            \ColorGradient{48.9}{63.6}{29.0}{my_awai_aoi}{48.9\%} & +3.0\%& 
            \ColorGradient{80.4}{85.8}{66.4}{my_awai_red}{80.4\%} & +9.8\%& 
            \ColorGradient{69.5}{80.0}{50.8}{my_awai_aoi}{69.5\%} & +9.1\% \\
            % Self: 
            Qwen$\times$2 & 
            \ColorGradient{54.8}{56.7}{45.3}{my_awai_red}{54.8\%} & +5.5\%& 
            \ColorGradient{58.6}{63.6}{29.0}{my_awai_aoi}{58.6\%} & +7.5\%& 
            \ColorGradient{80.1}{85.8}{66.4}{my_awai_red}{80.1\%} & +13.7\%& 
            \ColorGradient{66.4}{80.0}{50.8}{my_awai_aoi}{66.4\%} & +11.1\% \\
            % \arrayrulecolor{my_awai_gray}\bottomrule
            % \arrayrulecolor{my_awai_gray}
            \hline
            \textbf{Brainstorming(Ours)} \\
            Baichuan $\times$ Mistral & 
            \ColorGradient{52.5}{56.7}{45.3}{my_awai_red}{52.5\%} & +6.8\%& 
            \ColorGradient{47.8}{63.6}{29.0}{my_awai_aoi}{47.8\%} & +10.4\%& 
            \ColorGradient{80.3}{85.8}{66.4}{my_awai_red}{80.3\%} & +9.8\%& 
            \ColorGradient{66.1}{80.0}{50.8}{my_awai_aoi}{66.1\%} & +10.5\% \\
            Qwen $\times$ Baichuan & 
            \ColorGradient{53.2}{56.7}{45.3}{my_awai_red}{53.2\%} & +5.9\%& 
            \ColorGradient{54.1}{63.6}{29.0}{my_awai_aoi}{54.1\%} & \textbf{+14.1\%}& 
            \ColorGradient{81.4}{85.8}{66.4}{my_awai_red}{81.4\%} & +13.0\%& 
            \ColorGradient{64.0}{80.0}{50.8}{my_awai_aoi}{64.0\%} & +11.0\% \\
            Qwen $\times$ Mistral & 
            \ColorGradient{56.7}{56.7}{45.3}{my_awai_red}{56.7\%} & \textbf{+9.0\%}& 
            \ColorGradient{56.5}{63.6}{29.0}{my_awai_aoi}{56.5\%} & +8.0\%& 
            \ColorGradient{85.8}{85.8}{66.4}{my_awai_red}{85.8\%} & \textbf{+17.3\%}& 
            \ColorGradient{73.7}{80.0}{50.8}{my_awai_aoi}{73.7\%} & \textbf{+15.9\%} \\
          \end{tabular}
        }
        \end{center}
      \end{table}

% Tablet 2: gpt gemini vsSelf
\begin{table}[htbp]
        \begin{center}
          \caption{The accuracy of commercial models (Gemini-Pro and ChatGPT) using brainstorming, single model, ensemble and multi-agent reasoning strategies on different datasets.} 
          % Similar to results in Table \ref{tab:vs_baselines}, our method is also valid for popular online LLM services, which indicates the generalization ability of our method. }
          \label{tab:acc_commercial}
          % \begin{tabular}{cr|cccc|}
          %     & & \multicolumn{4}{c|}{Qwen-7B $\times$ Baichuan2-7B}\\
          %   & & MMLU & GSM & ARC-e & ARC-c \\
          %   \hline
            
          %   \multicolumn{2}{c|}{Game} & 53.2 & 54.1 & 81.4 & 64.0 \\ 
          %   \arrayrulecolor{my_awai_gray}\hline
          %   & Qwen-7B & 54.8 & 58.6 & 80.1 & 66.4 \\
          %   Self (2$\times$) & Baichuan2-7B & 45.9 & 32.4 & 72.8 & 53.3 \\
          %   & Mistral-7B &  &  &  &  \\
          %   % 3 & 23.113231 & c\\
          %   % 4 & 25.113231 & d\\
          % \end{tabular}
          \resizebox{0.75\textwidth}{!}{
          \begin{tabular}{c|cccccc}
              % & & \multicolumn{4}{c|}{Qwen-7B $\times$ Baichuan2-7B}\\
              \multirow{2}{*}{Method} &
              \multicolumn{2}{c}{MMLU} &
              % \makebox[0.1\textwidth][c]{MMLU(\%)} &
              \multicolumn{2}{c}{GSM} &
              % \makebox[0.1\textwidth][c]{GSM(\%)} &
              \multicolumn{2}{c}{ARC-c} \\
              % \makebox[0.1\textwidth][c]{ARC-c(\%)} \\
            % Method & MMLU(\%) & GSM(\%) & ARC-e(\%) & ARC-c(\%) \\
           & \makebox[0.7cm][c]{Acc.}& \makebox[0.3cm][c]{$\overline\Delta$}&
             \makebox[0.7cm][c]{Acc.}& \makebox[0.3cm][c]{$\overline\Delta$}& 
             \makebox[0.7cm][c]{Acc.}& \makebox[0.3cm][c]{$\overline\Delta$}\\
            \hline
            \textbf{Single-Model} \\
            GPT-3.5 & \ColorGradient{68.8}{71}{67.8}{my_awai_red}{68.8\%} & - &
            \ColorGradient{82.4}{85}{78.8}{my_awai_aoi}{82.4\%}& - & 
            \ColorGradient{84.6}{89}{84.6}{my_awai_red}{84.6\%}& -\\
            Gemini-Pro & \ColorGradient{68.0}{71}{67.8}{my_awai_red}{68.0\%}& - & 
            \ColorGradient{80.4}{85}{78.8}{my_awai_aoi}{80.4\%}& - & 
            \ColorGradient{86.3}{89}{84.6}{my_awai_red}{86.3\%}& - \\
            % \arrayrulecolor{my_awai_gray}
            % \hline
            % \textbf{Ensemble Vote} \\
            % GPT-3.5$\times$Gemini-Pro & 
            % \ColorGradient{68.4}{71}{67.8}{my_awai_red}{68.4\%}& +0.0\% & 
            % \ColorGradient{82.8}{85}{78.8}{my_awai_aoi}{82.8\%}& \textbf{+1.4\%} & 
            % \ColorGradient{85.1}{89}{84.6}{my_awai_red}{85.1\%}& -0.4\% \\
            \hline
            \textbf{Multi-Agent} \\
            GPT-3.5$\times$2 & \ColorGradient{70.9}{71}{67.8}{my_awai_red}{70.9\%}& +2.1\% & 
            \ColorGradient{83.2}{85}{78.8}{my_awai_aoi}{83.2\%}& +0.8\% & 
            \ColorGradient{86.0}{89}{84.6}{my_awai_red}{86.0\%}& +1.4\% \\
            Gemini-Pro$\times$2 &\ColorGradient{68.5}{71}{67.8}{my_awai_red}{68.5\%} & +0.5\% &
            \ColorGradient{78.8}{85}{78.8}{my_awai_aoi}{78.8\%}& -1.6\% & 
            \ColorGradient{86.9}{89}{84.6}{my_awai_red}{86.9\%}& +0.6\% \\
            % \arrayrulecolor{my_awai_gray}
            \hline
            \textbf{Brainstorming(Ours)} \\
            GPT-3.5 $\times$ Gemini-Pro  &\ColorGradient{71.0}{71}{67.8}{my_awai_red}{71.0\%} & \textbf{+2.6\%} &
            \ColorGradient{82.6}{85}{78.8}{my_awai_aoi}{82.6\%}& +1.2\% &
            \ColorGradient{88.4}{89}{84.6}{my_awai_red}{88.4\%}& \textbf{+3.0\%} \\
          \end{tabular}
          }
        \end{center}
      \end{table}
In Table \ref{tab:acc_open_source}, \ref{tab:acc_commercial}, we report the results of singel-model-based methods and multi-agent-based methods on MMLU, GSM, and ARC tasks. In each task, we observe that LLM brainstorming  overweighs the single model. 
In the fine-grained logical GSM dataset and factual reasoning ARC dataset, the brainstorming even achieve an impressive 10\% to 15\% improvement in average accuracy, shown as the Table \ref{tab:acc_open_source} $\overline{\triangle}$ column (Take brainstorming as an example, the $\overline{\triangle}$ is calculated by $Acc_{brain}(model_1,model_2) - \frac{Acc_{single}(model_1) + Acc_{single}(model_2)}{2}$). Although the ensemble model voting can result in an improvement in accuracy, the voting strategy still exhibits a lower reasoning accuracy than brainstorming on each task. This is due to the lack of interpretability of the voting strategy and its susceptibility to being misled by non-expert models.

% Compared with the multi-agent-based approaches, the reasoning accuracy of brainstorming is on average 
% 6.2\%, 4.8\% higher than that of multi-agent among the open source models in the logistic(GSM) and factual(ARC) datasets.
% 1.5\%, 6.2\%, 4.7\%, 4.9\%   
%3.4\%, 6.2\%, and 9.7\% 
Compared to the multi-agent based approach, the reasoning accuracy of brainstorming is on average 6.2\%, 4.8\% higher than multi-agent on logical (GSM) and factual (ARC) tasks respectively (shown as Table \ref{tab:acc_open_source}).
Although the multi-agents can produce better reasoning than a single model in most tasks, restricted by its basic model, the knowledge of different agents is overlapped, which leads to lower reasoning ability than brainstorming. Consequently, brainstorming by introducing different models and thus new knowledge can lead to correct answers on more tasks than multi-agent collaboration.

\makeatletter
\def\tkzKiviatGrad{\pgfutil@ifnextchar[{\tkz@KiviatGrad}{\tkz@KiviatGrad[]}} 
\def\tkz@KiviatGrad[#1](#2){% 
\begingroup
\pgfkeys{/kiviatgrad/.cd,
graduation distance= 0 pt,
prefix ={},
suffix={},
unity=1
 }
 \pgfqkeys{/kiviatgrad}{#1}% 
\let\tikz@label@distance@tmp\tikz@label@distance
\global\let\tikz@label@distance\tkz@kiv@grad
 \foreach \nv in {0,...,\tkz@kiv@lattice}{ %original: \foreach \nv in {1,...,\tkz@kiv@lattice}{
 \pgfmathparse{\tkz@kiv@unity*\nv} 
 \pgfmathtruncatemacro{\result}{\pgfmathresult+30} %original: \pgfmathtruncatemacro{\result}{\pgfmathresult}
 \protected@edef\tkz@kiv@gd{\tkz@kiv@prefix$\result$\tkz@kiv@suffix}
    \path[/kiviatgrad/.cd,#1] (0:0)--(360/\tkz@kiv@radial*#2:\nv*\tkz@kiv@gap) 
       node[label=(360/\tkz@kiv@radial*#2)-90:\tkz@kiv@gd] {}; 
      }
 \let\tikz@label@distance\tikz@label@distance@tmp  
\endgroup
}%
\makeatother

% \begin{figure}
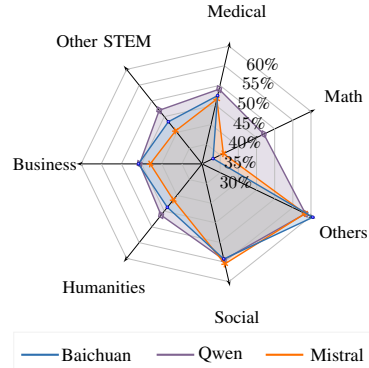
\begin{wrapfigure}[19]{r}{.4\textwidth}%靠文字内容的右侧
\centering
\resizebox{0.4\textwidth}{!}{
\begin{tikzpicture}[rotate=25.714,scale=.20]
% \tikzstyle{every node}=[font=\scriptsize]
\tkzKiviatDiagram[lattice = 6,
gap = 2,
step = 1,
label space = 3.6]%
{Math, Medical, Other STEM, Business, Humanities, Social, Others}
\tkzKiviatLine[thick,
               color = mypurple,
               mark = diamond,
               mark size = 10pt,
               fill = mypurple!40,
               ](3.404, 3.806, 3.404, 3.100, 3.257, 4.884, 5.696)
\tkzKiviatLine[thick,
               color = mistral,
               mark = star,
               mark size = 10pt,
               fill = mistral!40,
               mark size = 10pt](1.176, 3.301, 2.101, 2.571, 2.285, 5.094, 5.679)
\tkzKiviatLine[thick,
               color = my_awai_aoi,
               mark = ball,
               mark size = 10pt,
               fill = my_awai_aoi!40,
               mark size = 5pt](0.607, 3.478, 2.677, 3.151, 2.758, 4.844, 6.106)
\tkzKiviatGrad[prefix=\ ,unity=5, suffix=\%](1)
\end{tikzpicture}
}

\begin{tikzpicture}
\tikzstyle{every node}=[font=\scriptsize]
\draw[draw=black!05] (-0.1,-0.2) rectangle ++(4.8,0.5);
\draw [thick, my_awai_aoi] (0,0) -- (0.5,0); 
\node at (1.0,0) {Baichuan};
\draw [thick, mypurple] (1.8,0) -- (2.3,0); 
\node at (2.65,0) {Qwen};
\draw [thick, mistral] (3.25,0) -- (3.75,0); 
\node at (4.2,0) {Mistral};
\end{tikzpicture}

\caption{The accuracy distributions of the three open-source LLMs on different domains of the MMLU dataset. We divide the 57 tasks of MMLU into 7 major categories based on their fields.}
\label{fig:radar}
% \end{figure}
\end{wrapfigure}

Simultaneously, in order to explore the capability gains obtained by models with different capabilities during the brainstorming, we present the accuracy distributions of 7B models on each subdomains of the MMLU dataset.
Combining the results in Figure \ref{fig:radar} and Table \ref{tab:acc_open_source}, it can be observed that the performances of Qwen and Baichuan in the math subdomain are more different, and after brainstorming, the accuracy improvement they obtained on the GSM dataset is more pronounced compared to the other pairs. Similarly, Mistral and Qwen have large differences in the distribution of subdomains that require memorization, so their brainstorming can achieve better results on the ARC dataset, which emphasizes memorization.

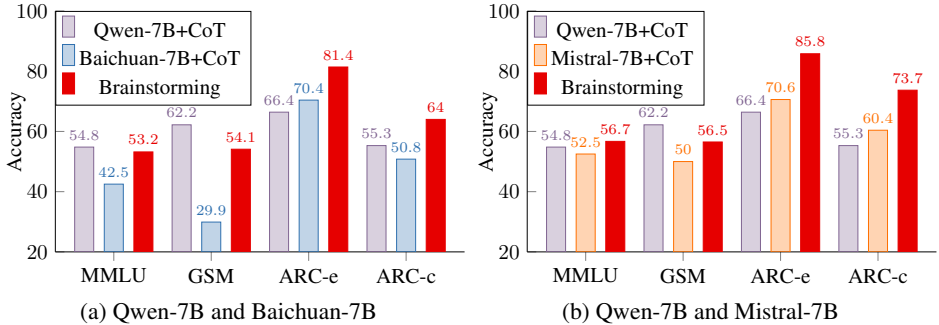
\begin{figure}[!ht]
        \centering
        \subfloat[Qwen-7B and Baichuan-7B]{
        \begin{tikzpicture}[scale=0.8]
        \pgfplotstableread[row sep=\\,col sep=&]{
        Dataset             & Qwen1.5-7B+CoT & Baichuan2-7B+CoT & Brainstorming \\
        MMLU                & 54.8 & 42.5  & 53.2  \\
        GSM   & 62.2 & 29.9 & 54.1  \\
        ARC-e            & 66.4 & 70.4 & 81.4 \\
        ARC-c       & 55.3 & 50.8 & 64.0 \\
        % 20--50   & 7.9  & 21.1 & 27.0 \\
        % 50+      & 3.0  & 22.3 & 28.6 \\
        }\vsCoT
        \begin{axis}
            [ybar,
             ylabel={Accuracy},
             ylabel shift = -4mm,
             bar width=9pt,
             ybar=5pt,
             width=0.6\textwidth,
             height=0.4\textwidth,
             legend style={at={(0,1)}, anchor=north west,legend columns=1},
             symbolic x coords={MMLU, GSM, ARC-e, ARC-c},
             xtick=data,
             enlarge x limits=0.2,
             axis y line*=left,
             axis x line*=bottom,
             nodes near coords,
             every node near coord/.append style={font=\scriptsize, fill opacity=1},
             nodes near coords align={vertical},
            ymin=20,ymax=100,
            legend image code/.code={\draw [#1] (0cm,-0.1cm) rectangle (0.2cm,0.25cm); }] 
            \addplot[
                mypurple,
                fill=mypurple,
                fill opacity=0.3,
                % pattern=north east lines, pattern color=mypurple,
            ] table[x=Dataset,y=Qwen1.5-7B+CoT]{\vsCoT};
            \addplot[my_awai_aoi,
                fill=my_awai_aoi,
                fill opacity=0.3,
                % pattern=north west lines, pattern color=my_awai_aoi,
            ] table[x=Dataset,y=Baichuan2-7B+CoT]{\vsCoT};
            \addplot[red!90!black,
                fill=red!90!black,
                % fill opacity=0.5,
                % pattern=crosshatch dots, pattern color=red,
            ] table[x=Dataset,y=Brainstorming]{\vsCoT};
            \legend{Qwen-7B+CoT, Baichuan-7B+CoT, Brainstorming} 
        \end{axis} 
        \end{tikzpicture}}
        \subfloat[Qwen-7B and Mistral-7B]{
        \begin{tikzpicture}[scale=0.8]
        \pgfplotstableread[row sep=\\,col sep=&]{
        Dataset  & Qwen1.5-7B+CoT & Mistral-7B+CoT & Brainstorming \\
        MMLU     & 54.8 & 52.5 & 56.7  \\
        GSM      & 62.2 & 50.0 & 56.5  \\
        ARC-e    & 66.4 & 70.6 & 85.8 \\
        ARC-c    & 55.3 & 60.4 & 73.7 \\
        % 20--50   & 7.9  & 21.1 & 27.0 \\
        % 50+      & 3.0  & 22.3 & 28.6 \\
        }\vsCoT
        \begin{axis}
            [ybar,
             ylabel={Accuracy},
             ylabel shift = -4mm,
             bar width=9pt,
             ybar=5pt,
             width=0.6\textwidth,
             height=0.4\textwidth,
             legend style={at={(0,1)}, anchor=north west,legend columns=1},
             symbolic x coords={MMLU, GSM, ARC-e, ARC-c},
             xtick=data,
             enlarge x limits=0.2,
             axis y line*=left,
             axis x line*=bottom,
             nodes near coords,
             every node near coord/.append style={font=\scriptsize, fill opacity=1},
             nodes near coords align={vertical},
            ymin=20,ymax=100,
            legend image code/.code={\draw [#1] (0cm,-0.1cm) rectangle (0.2cm,0.25cm); }] 
            \addplot[
                mypurple,
                fill=mypurple,
                fill opacity=0.3,
                % pattern=north east lines, pattern color=mypurple,
            ] table[x=Dataset,y=Qwen1.5-7B+CoT]{\vsCoT};
            \addplot[mistral,
                fill=mistral,
                fill opacity=0.3,
                % pattern=horizontal lines, pattern color=mistral,
            ] table[x=Dataset,y=Mistral-7B+CoT]{\vsCoT};
            \addplot[red!90!black,
                fill=red!90!black,
                % fill opacity=0.5,
                % pattern=crosshatch dots, pattern color=red,
            ] table[x=Dataset,y=Brainstorming]{\vsCoT};
            \legend{Qwen-7B+CoT, Mistral-7B+CoT, Brainstorming} 
        \end{axis} 
        \end{tikzpicture}}
        \caption{The accuracy comparison of brainstorming and CoT-base models on different datasets. CoT used in MMLU is 5-shot, 8-shot for GSM and 0-shot for ARC.}
        % {The models used for brainstorming are captioned below each chart. CoT used in MMLU is 5-shot, 8-shot for GSM and 0-shot for ARC. LLM Brainstorming can catch up with or even surpass few-shot CoT in the performance of across multiple tasks.}
        \label{fig:vs-cot}
    \end{figure}

\paragraph{Evaluation of Reasoning Efficiency. }
We show the accuracy of the multi-LLM brainstorming compared to a single model enhanced with CoT in Figure.\ref{fig:vs-cot}. CoT guides the model by demonstrating a small number of examples, leading to more accurate results. Instead of manual labeling, we use the reasoning process from other model as CoT-like example during the brainstorming. As can be seen from the figure, brainstorming achieves comparable reasoning accuracy to the CoT-based model across the different tasks. In particular, on the factual reasoning tasks (ARC-e and ARC-c datasets), brainstorming even outperforms the single LLM with CoT, improving the average accuracy by about 14.3\% compared to the CoT-based approaches. In the logical reasoning task (GSM), although Baichuan-7B as the base model (the average accuracy is around 29\%) is much less effective than the other models (the average accuracy is around 50\%), it still improves the reasoning accuracy to be comparable to that of the CoT-based method with the support of the brainstorming. This greatly reduces the cost of manual labeling, and provides a new solution for automation to improve the accuracy of model inference.

    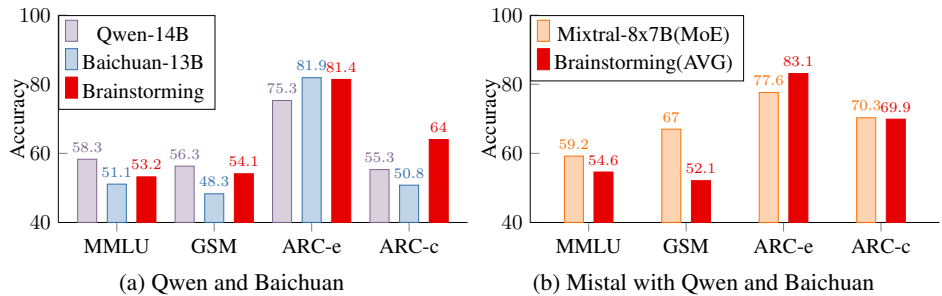
\begin{figure}[!ht]
        \centering
        \subfloat[Qwen and Baichuan]{
        \begin{tikzpicture}[scale=0.8]
        \pgfplotstableread[row sep=\\,col sep=&]{
        Dataset & Qwen1.5-14B & Baichuan2-13B & Brainstorming \\
        MMLU    & 58.3 & 51.1  & 53.2  \\
        GSM     & 56.3 & 48.3 & 54.1  \\
        ARC-e   & 75.3 & 81.9 & 81.4 \\
        ARC-c   & 55.3 & 50.8 & 64.0 \\
        }\vsjuuyon
        \begin{axis}
            [ybar,
             ylabel={Accuracy},
             ylabel shift = -4mm,
             bar width=9pt,
             ybar=5pt,
             width=0.6\textwidth,
             height=0.36\textwidth,
             legend style={at={(0,1)}, anchor=north west,legend columns=1},
             symbolic x coords={MMLU, GSM, ARC-e, ARC-c},
             xtick=data,
             enlarge x limits=0.2,
             axis y line*=left,
             axis x line*=bottom,
             nodes near coords,
             every node near coord/.append style={font=\scriptsize, fill opacity=1},
             nodes near coords align={vertical},
            ymin=40,ymax=100,
            legend image code/.code={\draw [#1] (0cm,-0.1cm) rectangle (0.2cm,0.25cm); }] 
            \addplot[
                mypurple,
                fill=mypurple,
                fill opacity=0.3,
                % pattern=north east lines, pattern color=mypurple,
            ] table[x=Dataset,y=Qwen1.5-14B]{\vsjuuyon};
            \addplot[my_awai_aoi,
                fill=my_awai_aoi,
                fill opacity=0.3,
                % pattern=north west lines, pattern color=my_awai_aoi,
            ] table[x=Dataset,y=Baichuan2-13B]{\vsjuuyon};
            \addplot[red!90!black,
                fill=red!90!black,
                % pattern=crosshatch dots, pattern color=red,
            ] table[x=Dataset,y=Brainstorming]{\vsjuuyon};
            \legend{Qwen-14B, Baichuan-13B, Brainstorming} 
        \end{axis} 
        \end{tikzpicture}}
        \subfloat[Mistal with Qwen and Baichuan]{
        \begin{tikzpicture}[scale=0.8]
        \pgfplotstableread[row sep=\\,col sep=&]{
        Dataset  & Mixtral-8x7B & Brainstorming \\
        MMLU     & 59.2 & 54.6  \\
        GSM      & 67.0 & 52.1  \\
        ARC-e    & 77.6 & 83.1 \\
        ARC-c    & 70.3 & 69.9 \\
        }\vsjuuyon
        \begin{axis}
            [ybar,
             ylabel={Accuracy},
             ylabel shift = -4mm,
             bar width=9pt,
             ybar=5pt,
             width=0.6\textwidth,
             height=0.36\textwidth,
             legend style={at={(0,1)}, anchor=north west,legend columns=1},
             symbolic x coords={MMLU, GSM, ARC-e, ARC-c},
             xtick=data,
             enlarge x limits=0.2,
             axis y line*=left,
             axis x line*=bottom,
             nodes near coords,
             every node near coord/.append style={font=\scriptsize, fill opacity=1},
             nodes near coords align={vertical},
            ymin=40,ymax=100,
            legend image code/.code={\draw [#1] (0cm,-0.1cm) rectangle (0.2cm,0.25cm); }] 
            \addplot[mistral,
                fill=mistral,
                fill opacity=0.3,
                % pattern=horizontal lines, pattern color=mistral,
            ] table[x=Dataset,y=Mixtral-8x7B]{\vsjuuyon};
            \addplot[red!90!black,
                fill=red!90!black,
                % pattern=crosshatch dots, pattern color=red,
            ] table[x=Dataset,y=Brainstorming]{\vsjuuyon};
            \legend{Mixtral-8x7B(MoE), Brainstorming(AVG)} 
        \end{axis} 
        \end{tikzpicture}}
    
        \caption{The accuracy comparison of the small-parameter models-based brainstorming (Qwen1.5-7B, Baichuan2-13B, Mistral-7B) with double-parameter models (Qwen1.5-14B, Baichuan2-13B) and with the MoE model (Mixtral 7 x 8B).}
        % \textbf{LLM Brainstorming Compared to LLM with around 14B parameters.} 0-shot for all the 4 datasets. Qwen1.5-7B and Baichuan2-7B are applied for Brainstorming in the left chart. The Brainstorming in the right chart is the average result of Qwen1.5-7B $\times$ Mistral-7B and Baichuan2-7B $\times$ Mistral-7B. The scale of active parameters of Mixtral-8x7B is around 12B, part of its total around 46B parameters.[cite moe] The LLM Brainstorming of two 7B models can almost be as one LLM with around 14B parameters on the reasoning performance.}
        \label{fig:vs14b}
    \end{figure}

\begin{table}[htbp]
    \begin{center}
    \caption{The accuracy and the time consumption under saving different number of historical rounds in the brainstorming between Qwen1.5-7B and Baichuan2-7B on GSM. The time consumption is the average run time for all questions in the GSM dataset. The compute resources is in Appendix \ref{apdx:compute-resources}.
    % The result is from the Brainstorming of Qwen1.5-7B and Baichuan2-7B on GSM. Saving Rounds=1,3,5,7 means that, besides the brainstorming of the first round (which contains the question and LLMs' first answers), we concatenate the latest 1,3,5,7 round(s) brainstorming dialogs and discard the previous as the input for the next round of brainstorming. Saving Rounds=7 means no discard. Our result shows that, 1) LLMs can summary the previous brainstorming in its next output; 2) partly slicing of brainstorming dialogs may lead to noise to LLMs; 3) No discard strategy can preserve the highest accuracy, however a long brainstorming dialog generated by this strategy results in a longer time consumption.
    }
    \label{tab:save_last_n_time}

    \begin{tabular}{c|cccc}
        \toprule 
        Saved Historical Rounds & 1 & 3 & 5 & 7 \\
        \hline
        Accuracy (\%) & 54.1 & 50.3 & 50.5 & 52.2 \\
        \hline
        Time Consumption (s) & 36.9 & 39.7 & 41.9 & 43.4 \\
        \bottomrule
    \end{tabular}
    % }
    \end{center}
\end{table}

We also demonstrate the accuracy comparison of brainstorming based approaches with the larger parameter model. Figure \ref{fig:vs14b} (a) illustrates that the reasoning accuracy of Qwen-7B and Baichuan-7B after brainstorming are able to reach an approximate level of the models with double-parameter (Qwen-14B and Baichuan-13B) on different tasks. Notice that, the brainstorming method is even 10\% more accurate than the two-parameter model on the ARC-c dataset, which requires more common knowledge to be memorized. Since Mistral does not have a 14B version of the model, we compare the results of Mistral-7B's participation in brainstorming with its MoE version (Mixtral 7$\times$8B) in Figure \ref{fig:vs14b} (b). We can see that the brainstorming-based approaches still possesses the ability to compete with the MoE model even when the scale of parameters is vastly different, and even obtains an outperform results of MoE in the ARC dataset. This indicates that brainstorming with models of different capabilities can dramatically improve the reasoning ability of the models through knowledge exchange. And many LLMs cannot be deployed on small devices due to the large number of parameters, while we found in our experiments that multiple small models can achieve similar capability to one large model after brainstorming, which provides a new solution for distributed deployment of large models.

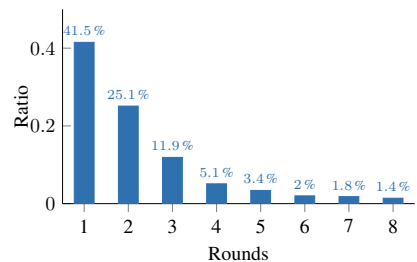
\begin{wrapfigure}[15]{r}{.4\textwidth}%靠文字内容的右侧

        \centering %图片居中
       
        \begin{tikzpicture}[scale=0.75]
        \pgfplotstableread[row sep=\\,col sep=&]{
        Rounds & GSM & MMLU \\ %& ARCE & ARCC
        1   & 0.247 & 0.415\\
        2   & 0.245 & 0.251  \\
        3   & 0.160 & 0.119\\
        4   & 0.111 & 0.051 \\
        5   & 0.056 & 0.034 \\
        6   & 0.046 & 0.020  \\
        7   & 0.030 & 0.018\\
        8   & 0.015 & 0.014\\
        }\convergeRounds
        \begin{axis}
            [ybar,
             xlabel={Rounds},
             ylabel={Ratio},
             ylabel shift = -2mm,
             bar width=10pt,
             ybar=3pt,
             width=0.56\textwidth,
             height=0.36\textwidth,
             legend style={at={(0,1)}, anchor=north west,legend columns=3},
             symbolic x coords={1, 2, 3, 4, 5, 6, 7, 8}, 
             xtick=data,
             enlarge x limits=0.07,
             axis y line*=left,
             axis x line*=bottom,
             nodes near coords={\pgfkeys{/pgf/fpu}\pgfmathparse{\pgfplotspointmeta*100}\pgfmathprintnumber{\pgfmathresult}\,\%
             },
             every node near coord/.append style={font=\scriptsize},
             nodes near coords align={vertical},
            ymin=0,ymax=0.5,
            legend image code/.code={\draw [#1] (0cm,-0.1cm) rectangle (0.2cm,0.25cm); }] 
    
            \addplot[my_awai_aoi,
                fill=my_awai_aoi,
            ] table[x=Rounds,y=MMLU]{\convergeRounds};
    
        \end{axis} 
        \end{tikzpicture}
       
        \caption{The proportion of tasks in the MMLU dataset where the brainstorming (Qwen1.5-7B and Baichuan2-7B) 
        reaches consensus 
        across rounds.}
        \label{fig:rounds_acc}  % 设置用于reference的label
    
    \end{wrapfigure}
\paragraph{Analysis of Model Properties. }
Considering that in a multi-round brainstorming scenario of the same problem, the reasoning process given by the model in a new round has a large portion of redundancy compared to the historical dialogs, so we evaluate whether to keep the historical dialogs or not. From Table \ref{tab:save_last_n_time}, we can see that the retention of more rounds of the reasoning process does not improve the final reasoning accuracy, but rather introduces additional noise to the model, which disturbs the thinking of the model, and thus leads to the decline of the final reasoning accuracy. Moreover, retaining more rounds of inference process not only increases the burden of parsing context for the model, but also brings more memory occupation and increase in time consumption. Therefore, we believe that with the LLM already having some summarization ability, retaining only the latest round of reasoning results in the brainstorming process can gain more benefits in terms of efficiency and reasoning accuracy.

\begin{figure}[ht] %插入图片
        \centering %图片居中
        % \resizebox{1\columnwidth}{!}{  %用于修改图片大小
        
    \subfloat[Qwen-7B $\times$ Baichuan-7B]{
            \begin{tikzpicture}[scale=0.8]
            % \scalefont{0.7} %设置字体大小
            \begin{axis}[
            sharp plot, %控制线的风格
            % title=Qwen1.5-7b \& Baichuan2-7B,%图像标题
            xmode=normal,% 控制坐标轴为线性
    %		ymode=log,% 控制坐标轴为对数
            xlabel=Rounds, %x坐标名
            ylabel=Accuracy, %y坐标名
            ylabel shift = -2mm,
            width=0.6\textwidth, height=0.4\textwidth,  %设置长和宽
            xmin=0,xmax=9,  % 设置x坐标范围
            ymin=20, ymax=80,  % 设置y坐标范围
            xtick={0,1,2,3,4,5,6,7,8}, %指定x轴刻度值。如果为空，则自动设置刻度线。即分割坐标轴
            % ytick={30,45,60,75}, %指定y轴刻度值。如果为空，则自动设置刻度线。即分割坐标轴
            xlabel near ticks, % 设置x坐标名位置靠近折线图
            ylabel near ticks, % 设置y坐标名位置靠近折线图
            ymajorgrids=true, % 启用/禁用 [公式] 轴上刻度线位置上的网格线
            grid style=dashed, % 设置网格线格式
            legend style={at={(1,0)},anchor=south east,legend columns=2}, % 设置标签位置
    %			legend columns=3, %设置标签列数
    %			legend pos=north west, % 设置折线对应标签的位置
    %			legend style={nodes={scale=0.6, transform shape}},  % 设置折线标签的格式
            ]
            
            %画第一条线，semithick设置线的粗细为0.6pt，mark是折线标示形状，options是mark形状的大小 ， olive!50!white是颜色，coordinates中包含要绘制的点的坐标
            \addplot+[semithick,mark=x,mark options={scale=0.8}, color=myorange] plot coordinates { 
                (1, 27.2)
    (2, 39.7)
    (3, 44.9)
    (4, 46.8)
    (5, 48.1)
    (6, 48.8)
    (7, 49.5)
    (8, 49.9)
            };
            \addlegendentry{MMLU}%第一条线标签
            
            %画第二条线
            \addplot+[semithick,mark options={scale=0.5}, color=my_awai_aoi] plot coordinates {
                (1, 21.2)
    (2, 35.0)
    (3, 41.5)
    (4, 46.5)
    (5, 48.1)
    (6, 49.5)
    (7, 50.3)
    (8, 51.0)
            };
            \addlegendentry{GSM} %第二条线标签
            
            % 画第三条线
            \addplot+[semithick,mark=diamond*, mark options={scale=0.8}, color=mypurple] plot coordinates {
                (1, 46.3)
    (2, 66.3)
    (3, 70.7)
    (4, 72.5)
    (5, 73.4)
    (6, 74.0)
    (7, 74.4)
    (8, 74.8)
            };
            \addlegendentry{ARC-e} %第三条线标签
    
            \addplot+[semithick,mark=triangle*,mark options={scale=0.8}, color=my_dark_midori] plot coordinates {
                (1, 32.4)
    (2, 50.1)
    (3, 55.0)
    (4, 56.8)
    (5, 57.8)
    (6, 58.9)
    (7, 59.2)
    (8, 59.7)
            };
            \addlegendentry{ARC-c} %第4条线标签
    \addplot [|<->|,smooth,thick,samples=50,red] coordinates {(1, 46.3) (1, 72.5)}node[xshift=-1mm,left,midway,pos=0.5,red,text width=5mm]{\scriptsize $\overline\Delta=26.2\%$};
    \addplot [|-|,smooth,thick,samples=50,red] coordinates {(4, 72.5) (4, 74.8)}node[yshift=2mm,left,midway,pos=0.1,red]{\scriptsize $\overline\Delta=2.3\%$};
    \addplot [-,dashed,thick,samples=50,red] coordinates {(1, 72.5) (4.1, 72.5)};%node[below,pos=.5,red]{};  
    \addplot [-,dashed,thick,samples=50,red] coordinates {(4, 74.8) (8.3, 74.8)};%node[below,pos=.5,red]{};  
            \end{axis}
            \end{tikzpicture}
            }
        % qw & mistral 
    \subfloat[Qwen-7B $\times$ Mistral-7B]{
        \begin{tikzpicture}[scale=0.8] %tikz图片
            % \scalefont{0.7} %设置字体大小
            \begin{axis}[
            sharp plot, %控制线的风格
            % title=Qwen1.5-7b \& Mistral-7B,%图像标题
            xmode=normal,% 控制坐标轴为线性
    %		ymode=log,% 控制坐标轴为对数
            xlabel=Rounds, %x坐标名
            ylabel=Accuracy, %y坐标名
            ylabel shift = -2mm,
            width=0.6\textwidth, height=0.4\textwidth,  %设置长和宽
            xmin=0,xmax=9,  % 设置x坐标范围
            ymin=25, ymax=85,  % 设置y坐标范围
            xtick={0,1,2,3,4,5,6,7,8}, %指定x轴刻度值。如果为空，则自动设置刻度线。即分割坐标轴
            % ytick={30,45,60,75}, %指定y轴刻度值。如果为空，则自动设置刻度线。即分割坐标轴
            xlabel near ticks, % 设置x坐标名位置靠近折线图
            ylabel near ticks, % 设置y坐标名位置靠近折线图
            ymajorgrids=true, % 启用/禁用 [公式] 轴上刻度线位置上的网格线
            grid style=dashed, % 设置网格线格式
            legend style={at={(1,0)},anchor=south east,legend columns=2}, % 设置标签位置
    %			legend columns=3, %设置标签列数
    %			legend pos=north west, % 设置折线对应标签的位置
    %			legend style={nodes={scale=0.6, transform shape}},  % 设置折线标签的格式
            ]
            
            %画第一条线，semithick设置线的粗细为0.6pt，mark是折线标示形状，options是mark形状的大小 ， olive!50!white是颜色，coordinates中包含要绘制的点的坐标
            \addplot+[semithick,mark=x,mark options={scale=0.8}, color=myorange] plot coordinates { 
                (1, 28.6)
    (2, 43.2)
    (3, 48.6)
    (4, 51.2)
    (5, 52.7)
    (6, 53.8)
    (7, 54.4)
    (8, 54.8)
            };
            \addlegendentry{MMLU}%第一条线标签
            
            %画第二条线
            \addplot+[semithick,mark options={scale=0.5}, color=my_awai_aoi] plot coordinates {
                (1, 29.9)
    (2, 43.4)
    (3, 47.5)
    (4, 49.7)
    (5, 51.3)
    (6, 52.3)
    (7, 53.1)
    (8, 53.5)
            };
            \addlegendentry{GSM} %第二条线标签
            
            % 画第三条线
            \addplot+[semithick,mark=diamond*, mark options={scale=0.8}, color=mypurple] plot coordinates {
                (1, 47.9)
    (2, 73.5)
    (3, 78.0)
    (4, 79.7)
    (5, 80.1)
    (6, 80.4)
    (7, 80.7)
    (8, 80.7)
            };
            \addlegendentry{ARC-e} %第三条线标签
    
            \addplot+[semithick,mark=triangle*,mark options={scale=0.8}, color=my_dark_midori] plot coordinates {
                (1, 34.6)
    (2, 60.2)
    (3, 64.2)
    (4, 66.2)
    (5, 67.6)
    (6, 68.4)
    (7, 69.2)
    (8, 69.5)
            };
            \addlegendentry{ARC-c} %第4条线标签
    \addplot [|<->|,smooth,thick,samples=50,red] coordinates {(1, 47.9) (1, 79.7)}node[xshift=-1mm,left,midway,pos=0.55,red,text width=5mm]{\scriptsize $\overline\Delta=31.8\%$};
    \addplot [|-|,smooth,thick,samples=50,red] coordinates {(4, 79.7) (4, 80.7)}node[yshift=2mm,left,midway,pos=0.1,red]{\scriptsize $\overline\Delta=1.0\%$};
    \addplot [-,dashed,thick,samples=50,red] coordinates {(1, 79.7) (4.1, 79.7)};%node[below,pos=.5,red]{};  
    \addplot [-,dashed,thick,samples=50,red] coordinates {(4, 80.7) (8.3, 80.7)};%node[below,pos=.5,red]{};  
    
            \end{axis}
            \end{tikzpicture}
            }
        \caption{The accuracy of multi-model brainstorming at different number of dialog rounds, with an upper limit of 8 rounds. The red bidirectional arrows mark the change in accuracy of Brainstorming for 1 to 4 and 4 to 8 rounds with the ARC-e dataset.}
        \label{fig:diff_round}  % 设置用于reference的label
    \end{figure}
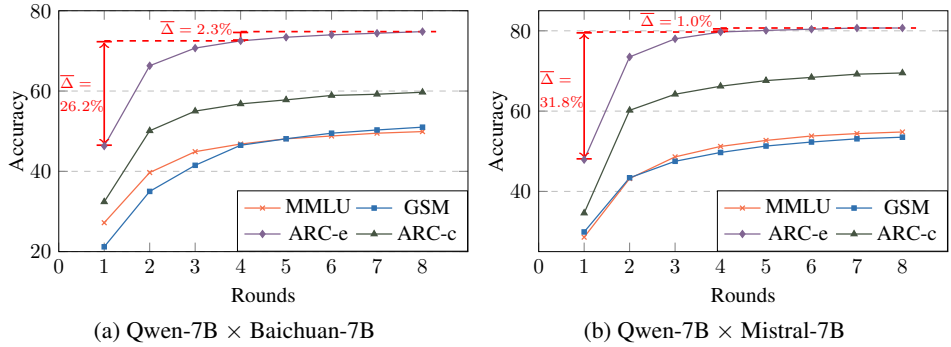
We also explored the number of rounds required for brainstorming. As can be seen in Figure \ref{fig:rounds_acc}, the vast majority of questions can be answered consistently with about 4 rounds, with only a few questions requiring more rounds of discussion. It can also be seen in Figure \ref{fig:diff_round} that the improvement in accuracy from single model reasoning to four rounds of interaction is the most significant, up to 30\%. As the number of rounds rises, the gains from interaction gradually reduce, and the reasoning accuracy tends to converge when the number of rounds reaches about 4, and the accuracy increases by only 2\% from 4 rounds to 8 rounds. So the brainstorming %only 
requires fewer rounds to obtain a substantial increase in reasoning accuracy.

\begin{figure*}[htbp]
    \centering
    \includegraphics[width=\textwidth]{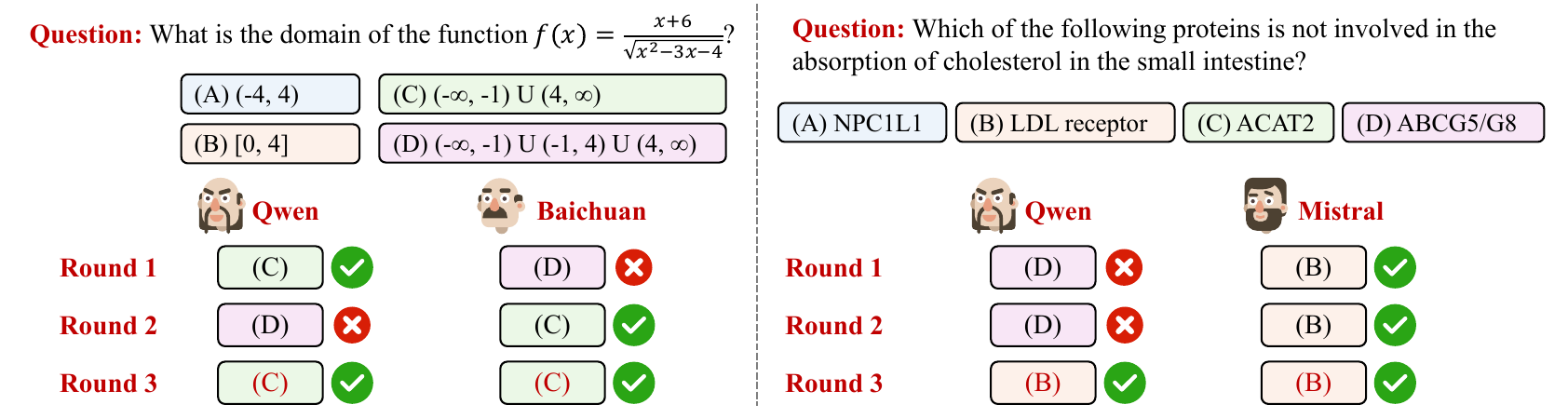}
    \caption{Changes in the reasoning results of Qwen-7B and Mistral-7B during brainstorming over logic and fact tasks.}
    \label{fig:normal_logic_factial_task}
\end{figure*}

\begin{figure*}[!ht]
    \centering
    \includegraphics[width=1\textwidth]{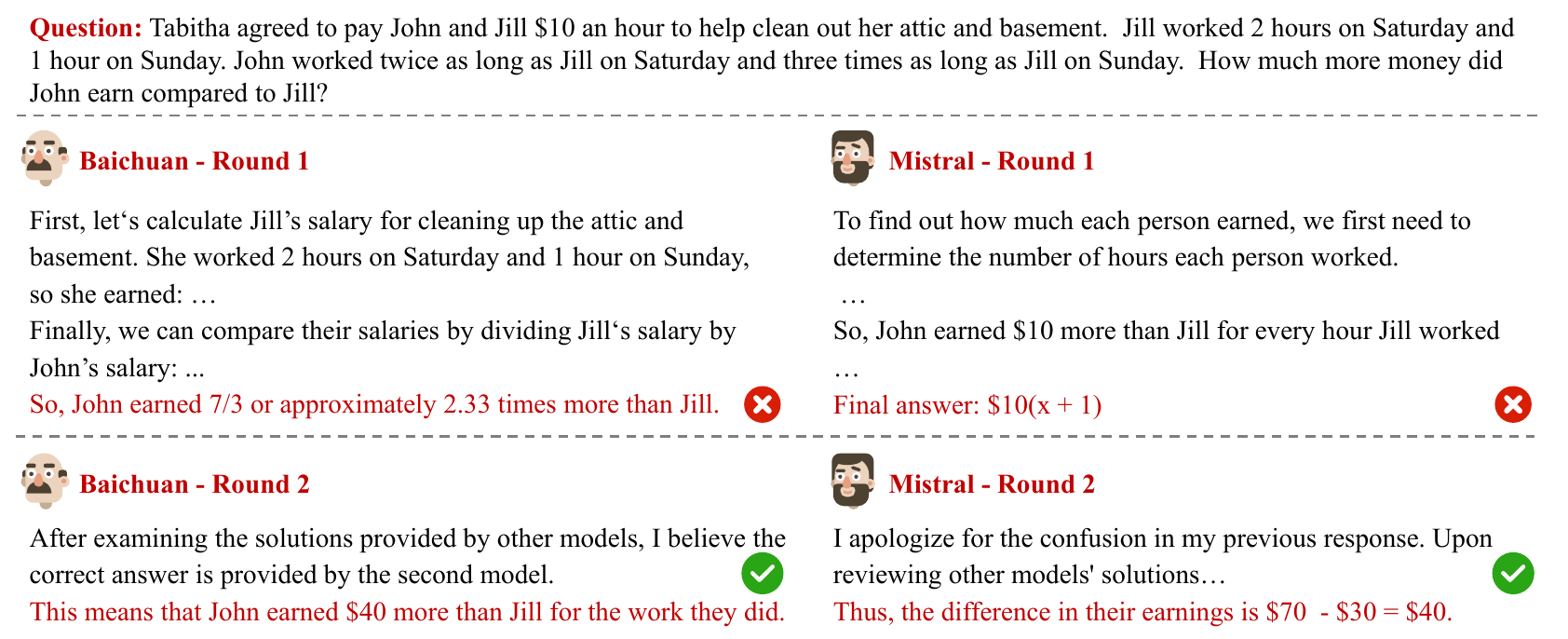}
    \caption{An illustrative example from Baichuan and Mistral brainstorming reasoning processes on a challenging mathematical task.}
    \label{fig:challenging_logic_task}
\end{figure*}

\paragraph{Case Study. }
In Figure \ref{fig:normal_logic_factial_task}, we show the process of brainstorming among multiple models in a mathematical reasoning task and a biological memory task. We found that even though the models do not arrive at a unified answer in their initial reasoning, after multiple rounds of interaction, the initially incorrect model can correct its reasoning based on the reasoning process of other models, thus arriving at a correct answer that is consistent across multiple models.

\begin{figure*}[htbp]
    \centering
    \includegraphics[width=\textwidth]{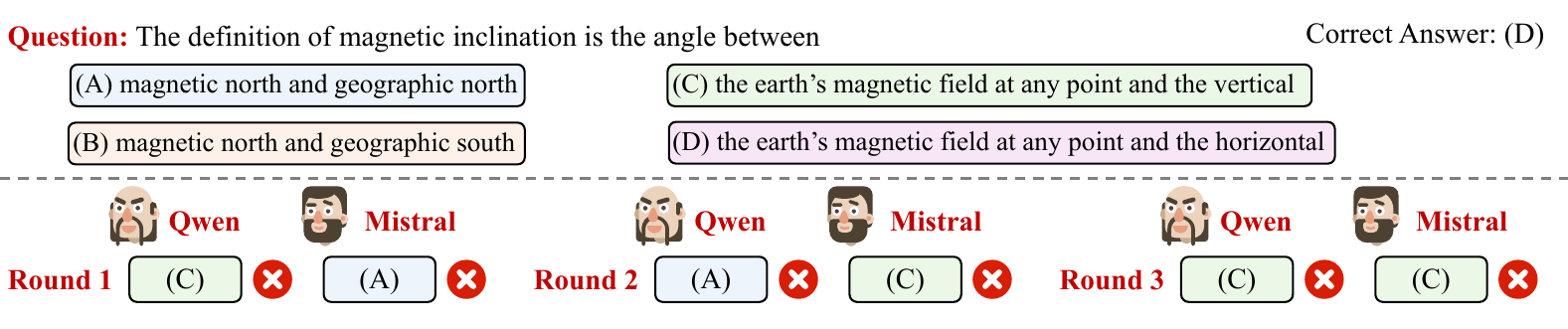}
    \caption{Qwen and Mistral's brainstorming results without domain knowledge.}
    \label{fig:factual_task_without_knowledge}
\end{figure*}

In some more challenging tasks, as illustrated in Figure \ref{fig:challenging_logic_task}, neither model is able to provide the correct answer initially. Following a series of brainstorming, the models are able to reach a consensus of the correct answer. In the case of the factual task in Figure \ref{fig:factual_task_without_knowledge} that relies on memory, the models lack  requisite knowledge and unable to provide the correct answer even after exchanging information with each other for several rounds. This is consistent with the typical human discussion process without requisite knowledge.

\begin{figure*}[!t]
    \centering
    \includegraphics[width=\textwidth]{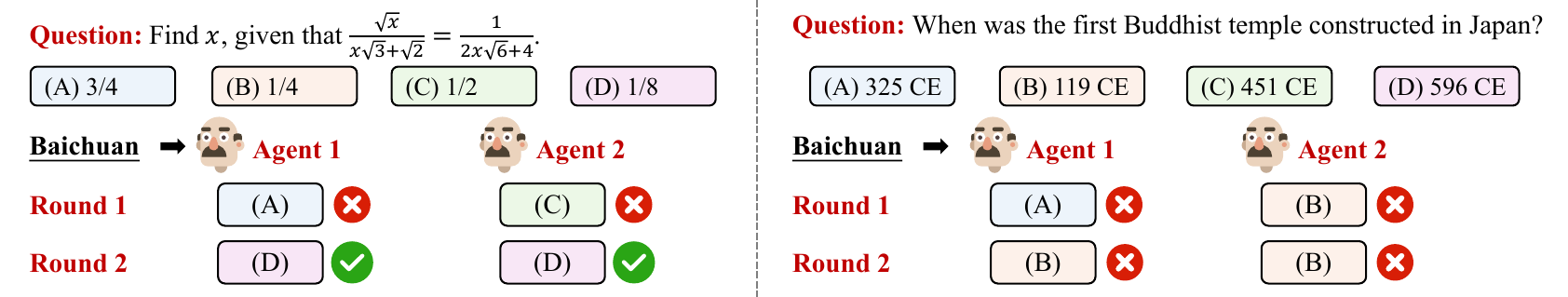}
    \caption{Results of debates between multiple agents with the same knowledge generated by Baichuan in logical and factual tasks.}
    \label{fig:self_logic_factual_task}
\end{figure*}

\begin{figure*}[!t]
    \centering
    \includegraphics[width=0.9\textwidth]{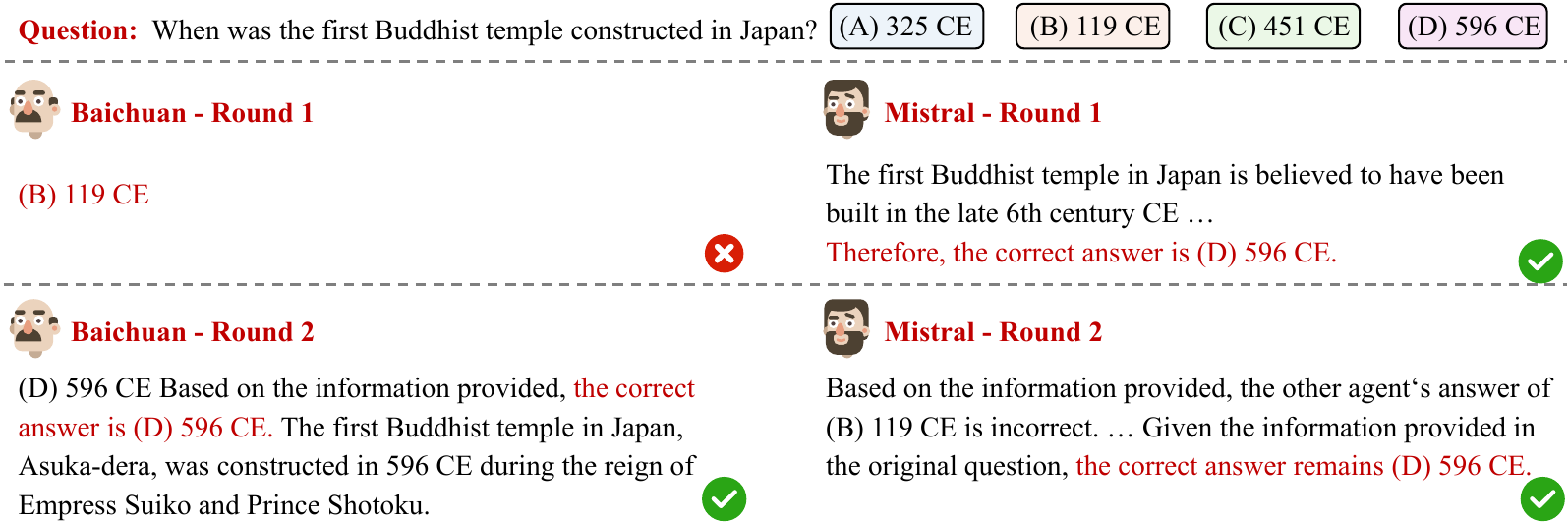}
    \caption{An illustrative of Baichuan and Mistral brainstorming when answering the same historical factual question as that in the right-hand side of Figure \ref{fig:self_logic_factual_task}.}
    \label{fig:factual_task_with_knowledge}
\end{figure*}

This phenomenon is further analyzed in the context of multi-agent debate and multi-model brainstorming. Figure \ref{fig:self_logic_factual_task} illustrates that, despite the single agent can not provide a correct answer in a logic task (left question of Figure \ref{fig:self_logic_factual_task}), multi-agent is able to arrive at the correct answer through a debate. However, when a multi-agent debate is introduced for the same factual task (right question of Figure \ref{fig:self_logic_factual_task}), it is still struggle to obtain the correct answer when each agent gives a wrong answer. This also demonstrates that the agents cannot obtain non-existent knowledge through interaction. However, by introducing another model with different capability, as shown in Figure \ref{fig:factual_task_with_knowledge}, the knowledge could be introduced into the thinking process from the other model, and thus the wrong answer can be corrected after brainstorming. The detail brainstorming dialog cases are in Appendix \ref{apdx:examples}.

\section{Discussion and Limitations}
In this paper, we propose the multi-model brainstorming which contributes to improving the kownledge reasoning. We emphasize the necessity of the diverse capability and propose distinct interaction strategies as integral components within the brainstorming. Our quantitative analysis of the reasoning results conveys intuitions about knowledge exchange by brainstorming, and substantiates the effectiveness of our approach in enhancing knowledge reasoning and efficiency.

\textbf{Limitations. }In comparison to other prompting techniques, our multi-model brainstorming procedure is more computationally expensive because it requires both multiple generations and more computational space. Nevertheless, we believe that the intermediate discussion process and the generated consensus result can be distilled back to fine-tune the base model and further enhance the original knowledge reasoning ability of LLMs.

% \newpage
\bibliographystyle{plain}
\bibliography{neurips_2024}
%%%%%%%%%%%%%%%%%%%%%%%%%%%%%%%%%%%%%%%%%%%%%%%%%%%%%%%%%%%%

\end{document}